\documentclass[runningheads]{llncs}
\usepackage[T1]{fontenc}
\usepackage{graphicx,verbatim}
\usepackage[colorlinks=true,citecolor=blue,urlcolor=blue]{hyperref}
\usepackage{amsmath}
\usepackage{amssymb}
\usepackage{caption}
\usepackage{booktabs}
\usepackage{multirow}
\usepackage{bbding}
\begin{document}
%
\title{Bridging Vision Foundation Model Priors with \\ CLIP for Spatial-aware Few-shot  Anomaly \\Detection in Medical Images}
\titlerunning{Bridging Vision Foundation Model Priors with CLIP}
%
\author{Juzheng Miao\inst{1} \and Yuchen Yuan\inst{1} \and Cheng Chen\inst{2,3}\textsuperscript{(}\Envelope\textsuperscript{)} \and Pheng-Ann Heng\inst{1,4}}

\authorrunning{J. Miao et al.}

\institute{Department of Computer Science and Engineering,\\The Chinese University of Hong Kong, Hong Kong, China \and Department of Electrical and Computer Engineering,\\The University of Hong Kong, Hong Kong, China \and School of Biomedical Engineering, The University of Hong Kong, Hong Kong, China \\  \email{cchen@eee.hku.hk} \and Institute of Medical Intelligence and XR,\\The Chinese University of Hong Kong, Hong Kong, China
}
\renewcommand{\thefootnote}{}
\footnotetext{J. Miao and Y. Yuan—contributed equally.}
  
\maketitle              
\begin{abstract}
Vision-Language Models such as CLIP enable effective few-shot medical anomaly detection (AD) via strong image-text semantic alignment. However, their globally contrastive pretraining lacks explicit spatial supervision, limiting precise lesion localization.
In contrast, Vision Foundation Models (VFMs) such as DINO learn spatially coherent patch representations via self-distillation and local-to-global consistency, better capturing fine-grained anatomical structures.
Leveraging this complementarity, we propose Spatial-FAD, a spatial-aware few-shot medical AD framework that improves lesion localization by combining VFM spatial priors with CLIP semantics.
Specifically, we introduce a VFM-enhanced adapter that injects a structural affinity prior derived from DINO into CLIP features. This structure-guided refinement encourages visual embeddings to better adhere to lesion boundaries while maintaining semantic alignment. 
%
To address the loss of spatial detail from patchification and the limited input resolution of CLIP, we adopt a sliding-window aggregation strategy. This generates high-resolution, spatially dense embeddings to further enhance localization granularity.
Moreover, we introduce a prototype-enhanced support memory scheme to efficiently exploit the few-shot support set. This module stores compact prototypes for normal and abnormal patterns, reducing memory costs while boosting performance by fusing patch-to-prototype and image-text similarities.
%
Extensive experiments on three benchmark datasets, including Liver CT, Retinal OCT, and Brain MRI, demonstrate that Spatial-FAD significantly outperforms state-of-the-art methods, {especially in lesion segmentation}. Notably, in the 4-shot scenario, our method achieves an average improvement of over 11.4$\%$ in Dice score and 1.8$\%$ in AUC.
Code is available at: \url{https://github.com/JuzhengMiao/Spatial-FAD}.

\end{abstract}
\section{Introduction}

Medical anomaly detection, aiming to not only distinguish between normal images and abnormal images with lesions but also segment the spatial lesion locations, plays an important role in assisting decision-making and facilitating early interventions in clinical practice~\cite{fernando2021deep,su2021few,zhang2020viral}. Previous mainstream methods tend to leverage a large scale of normal samples to model the normal distribution and treat data that deviates from this distribution as anomalies~\cite{deng2022anomaly,salehi2021multiresolution}, resulting in extremely expensive data collection costs.

Recent vision-language models (VLMs) pretrained on large-scale image-text pairs, e.g., CLIP~\cite{radford2021learning}, have shown strong few-shot anomaly detection performance using only a handful of normal and abnormal images, enabling more data-efficient anomaly detection~\cite{huang2024adapting,zhang2024mediclip,zhou2025ultraad}. In this paradigm, a vision encoder extracts image patch embeddings while a text encoder produces prompt embeddings, and anomaly scores are then derived from cosine similarities between patch and text features to form a localization map. To transfer the CLIP-based anomaly detection framework to medical domains, Huang et al.\cite{huang2024adapting} insert residual adapters into CLIP’s vision encoder to reduce the gap between natural and medical images. In parallel, learnable prompts replace handcrafted text templates to better separate normal and abnormal semantics\cite{li2025kanoclip,ma2025aligning,shiri2025madclip}. MediCLIP further boosts few-shot performance by synthesizing diverse anomaly patterns to mimic common diseases in medical images~\cite{zhang2024mediclip}.

Although these VLM-based methods achieve better AUC scores, their lesion segmentation quality is often unsatisfactory. As shown in Fig.~\ref{visual}, the predicted maps are poorly aligned with lesion boundaries and sometimes even miss the correct locations compared with the ground-truth masks. This limitation largely stems from CLIP’s image-level contrastive pretraining that aligns global image features with text, but does not explicitly enforce fine-grained correspondence between local patch features and lesion-specific descriptions~\cite{wang2024sclip}.
Moreover, many approaches perform inference mainly by comparing test images with a set of predefined prompts~\cite{shiri2025madclip,zhang2024mediclip}, underutilizing the information contained in the few-shot training examples. 
MVFA~\cite{huang2024adapting} partially addresses this by storing normal training images in a memory bank and using support-query similarity as an auxiliary signal. 
However, it ignores lesion patterns in abnormal support images and stores dense patch-level embeddings for every support image, leading to substantial memory cost and limited scalability as the number of shots increases.

In this paper, we propose a novel paradigm \textbf{Spatial-FAD} that bridges Vision Foundation Model (VFM) priors with CLIP for spatial-aware few-shot anomaly detection in medical images, with a particular focus on improving the spatial accuracy of lesion segmentation. Our approach is motivated by the observation that, while CLIP provides strong semantic alignment, self-supervised VFMs such as DINO~\cite{caron2021emerging} capture local spatial structure more effectively. In contrast to CLIP’s image-level contrastive objective, DINO is trained via self-distillation with local-to-global correspondence, which encourages patch embeddings to be spatially coherent and to align with object boundaries~\cite{caron2021emerging,liumatcher}.
Building on this, we introduce a VFM-enhanced adapter that combines DINO’s spatial precision in structure with CLIP’s semantic richness. Specifically, we compute a DINO-based self-attention map from patch embeddings and use it as a structural affinity prior that encodes patch-wise correlations, indicating regions likely belonging to the same anatomy or lesion. 
This affinity is then applied to CLIP patch features to perform spatial prior-guided refinement to obtain embeddings that remain well aligned with text prompts while better adhering to lesion boundaries. Lightweight adapters are subsequently used to further adapt these features to the target medical domain.
To further improve localization accuracy, we additionally adopt a sliding-window aggregation strategy to enhance the feature resolution.
Considering the problem of spatial information loss caused by the transformer patchification and the small input size, our proposed method efficiently produces spatially dense embeddings that preserve fine detail by extracting and aggregating features over overlapping windows at the high-resolution scale.
Finally, we propose a prototype-enhanced support memory scheme to better exploit the few-shot support set. Instead of storing all patch embeddings, we compute compact prototypes from both normal and abnormal examples, substantially reducing memory and computation.
Then, during inference, patch-to-prototype similarities are fused with text-based similarity scores for the final prediction.
Extensive experiments on three benchmark datasets of few-shot anomaly detection in medical images, including liver tumor in CT, retinal edema in OCT, and brain tumor in MRI, demonstrate consistent gains, especially in anomaly segmentation. Specifically, in the 4-shot scenario, it achieves an average improvement of over 11.4\% in Dice score and more than 1.8\% in AUC across all three datasets, with a remarkable gain of 23.74\% Dice and 9.42\% AUC on the LiverCT dataset.

\section{Method}
\begin{figure}[!t]
 \centering
 \includegraphics[width=0.95\textwidth]{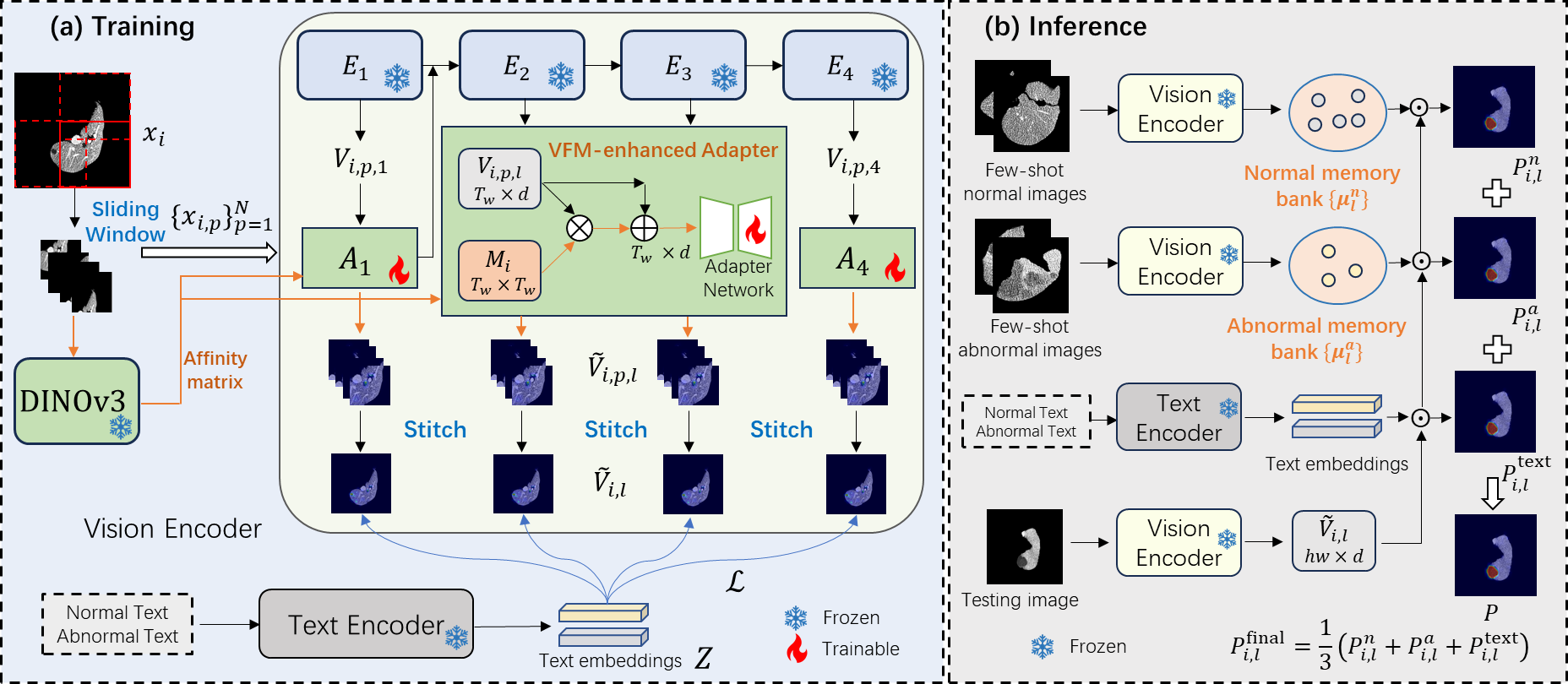}
 \caption{The overview of our proposed method. The VFM-enhanced adapter equipped with a sliding-window strategy is proposed to improve the spatial accuracy of lesion segmentation. During inference, a prototype-enhanced support memory scheme is developed to fully utilize the few-shot training examples.} \label{fig:method}
\end{figure}

Fig.~\ref{fig:method} provides an overview of our \textbf{Spatial-FAD} framework.
Our approach addresses the critical limitation of VLM-based methods for few-shot medical anomaly detection: while CLIP yields strong image-level discrimination, its poor patch-text alignment often results in coarse and fragmented pixel-level similarity maps, leading to low spatial accuracy. 
To overcome this, we introduce two key components to enhance the spatial localization: (i) a \textbf{VFM-enhanced adapter} that injects a structural affinity prior derived from DINO's patch embeddings into CLIP features, and (ii) a \textbf{sliding-window aggregation} strategy that aggregates spatially dense features from the high-resolution image in a computationally efficient way, avoiding the information loss of patchification downsampling.
Furthermore, to fully exploit the few-shot training support set at the inference phase, a \textbf{prototype-enhanced support memory} that constructs compact prototypes for both normal and abnormal patterns is proposed.

\subsection{Problem Formulation and Baseline Architecture}
\emph{\(k\)-shot} medical anomaly detection aims to identify anomalies using a small labeled support set
\(
\mathcal{D}=\{(x_i,c_i,Y_i)\}_{i=1}^{2k},
\)
where \(x_i\) is the input image, \(c_i\in\{0,1\}\) is the image-level label (0: normal, 1: abnormal), and 
\(Y_i\in\{0,1\}^{H\times W}\) is the pixel-level lesion mask.
Following MVFA~\cite{huang2024adapting}, we build our detector on CLIP, inserting lightweight learnable adapters \(\{A_l\}_{l=1}^{4}\) into four stages of the vision encoder \(\{E_l\}\). 
Given \(x_i\), stage \(l\) outputs visual patch embeddings \(V_{i,l}\in\mathbb{R}^{T\times d}\), where \(T=h\times w\) is the number of patches and $(h,w)$ denotes spatial grid size after patchification.
Adapters update these embeddings to \(\tilde{V}_{i,l}=A_l(V_{i,l})\).
Simultaneously, we obtain text embeddings \(Z\in\mathbb{R}^{2\times d}\) for normal and abnormal prompts from the CLIP text encoder. 
The vision-to-text similarity is computed via cosine similarity: $
S^{\text{text}}_{i,l}=\cos(\tilde{V}_{i,l},Z)\in\mathbb{R}^{T\times 2}.
$
Then, we reshape and upsample \(S^{\text{text}}_{i,l}\) to the image resolution to obtain the prediction map \(\textbf{P}_{i,l}\in\mathbb{R}^{H\times W\times 2}\).
For image-level classification, we obtain the anomaly score by applying channel-wise softmax, selecting the abnormal channel to yield the text-based anomaly score \(P^{\text{text}}_{i,l}\in\mathbb{R}^{H\times W}\), and subsequently applying global max pooling.
The total training objective sums the per-stage losses, combining pixel-level Dice loss and Focal loss and image-level classification loss via binary cross-entropy (BCE):
\begin{equation}
\label{eq:loss}
\mathcal{L}=\sum_{l=1}^{4} \left[ \mathcal{L}_{\text{dice}}(\textbf{P}_{i,l},{Y}_i)
+\mathcal{L}_{\text{focal}}(\textbf{P}_{i,l},{Y}_i)
+\mathcal{L}_{\text{bce}}\!\left(\max_{h,w} {P}_{i,l}^{\text{text}}(h,w),\,c_i\right) \right].
\end{equation}

\subsection{VFM-enhanced Structural Affinity with Sliding-window Aggregation}
\label{sec:structural_bias}
Standard CLIP training lacks explicit constraints for fine-grained patch correspondence, which often yields noisy boundaries and fragmented responses in pixel-level similarity maps. 
To enhance the performance of spatial-aware anomaly detection, we introduce two complementary components: 

\textbf{VFM-enhanced adapter with structural affinity prior (VFMEA).}
Unlike CLIP, self-supervised VFMs like DINO learn spatially coherent representations that naturally delineate object boundaries. 
To fully utilize this property, we compute a self-attention map based on DINO's patch embeddings to serve as a structural affinity prior.
Given an image \(x_i\), we extract DINOv3~\cite{simeoni2025dinov3} patch embeddings \(D_i\in\mathbb{R}^{T\times d_v}\) and compute a patch-to-patch affinity matrix: $
\label{eq:affinity}
M_i=\text{softmax}\!\left(\bar{D}_i\bar{D}_i^{\top}\right)\in\mathbb{R}^{T\times T},
$
where \(\bar{D}_i\) denotes normalized features. 
\(M_i\) encodes the likelihood of patches belonging to the same anatomical structure.
The affinity prior is then applied to CLIP features before the adapter at each $E_l$:
\begin{equation}
\label{eq:refine}
V'_{i,l}=\lambda V_{i,l}+(1-\lambda)\,M_i V_{i,l},
\end{equation}
where \(\lambda\in[0,1]\) balances the original semantic features and the structural guidance. The refined features \(V'_{i,l}\) are then fed into the learnable adapter to obtain the final visual embeddings $\tilde{V}_{i,l}=A_l(V'_{i,l})$. This injection of local structural prior effectively smooths the feature map, reducing fragmented responses and aligning predictions with anatomical boundaries.

\textbf{Sliding-window Aggregation (SWA).}
While the affinity prior improves boundary definition, the detection of small lesions is still constrained by the low input resolution of standard CLIP due to the use of patchification downsampling and the small input size.
To further enhance the spatial accuracy of lesion detection, a sliding-window aggregation strategy is introduced. 
Instead of resizing the entire image to the limited input size of CLIP, which causes information loss for details, this strategy traverses the image at a larger scale and aggregates features from local windows for a high-resolution feature map.
Given a high-resolution image \(x_{i}\in\mathbb{R}^{H\times W}\) (e.g., \(512\times512\)), we generate \(N\) overlapping crops \(\{x_{i,p}\}_{p=1}^{N}\) of size \(s\times s\) (e.g., \(s=336\) for CLIP-L).
For each crop, we run the same CLIP vision encoder and stage-\(l\) adapter \(A_l\) to obtain adapted window-level patch embeddings:
$
\tilde{V}_{i,p,l}=A_l\!\left(V_{i,p,l}\right)\in\mathbb{R}^{T_w\times d},
$
where \(V_{i,p,l}\) denotes the stage-\(l\) patch embeddings extracted from the crop image \(x_{i,p}\), and \(T_w\) is the number of patches per window.
To reduce the computational cost, $N$ crops are concatenated in the batch dimension, allowing for obtaining the output embeddings in one forward pass.
We then stitch these adapted features back onto the full-image canvas to form a spatially dense feature map: $
\tilde{V}_{i,l}=\text{Stitch}\big(\{\tilde{V}_{i,p,l}\}_{p=1}^{N}\big)\in\mathbb{R}^{T\times d}$
where \(\text{Stitch}(\cdot)\) places each window feature grid at its corresponding spatial location and aggregates overlapping regions by averaging.
This strategy preserves fine-grained details from the original image and improves the localization of small anomalies without incurring prohibitive computational time.

\subsection{Prototype-enhanced Support Memory at Test Time}
Existing methods often under-utilize the support set, either ignoring it during inference or storing memory-intensive dense patch embeddings~\cite{huang2024adapting}.
To enhance the test-time performance with the few-shot support set, we propose a prototype-enhanced support memory (PSM) scheme, functioning as a more efficient memory approach by constructing compact prototypes for both normal and abnormal classes.
Specifically, for each stage \(l\), we collect the adapter-refined patch embeddings from the support set and separate them by image label:
\(
\mathcal{V}^{n}_l=\{ \tilde{V}_{i,l}\mid c_i=0\},\;
\mathcal{V}^{a}_l=\{ \tilde{V}_{i,l}\mid c_i=1\}.
\)
K-means algorithm~\cite{ahmed2020k} is then conducted on the collected patch embeddings to obtain \(n_n\) normal prototypes
\(\boldsymbol{\mu}^{n}_l=\{\mu^{n}_{l,j}\}_{j=1}^{n_n}\) and \(n_a\) abnormal prototypes
\(\boldsymbol{\mu}^{a}_l=\{\mu^{a}_{l,j}\}_{j=1}^{n_a}\).
During inference, we compute the similarity between test image features \(\tilde{V}_{i,l}\) and these prototypes, reducing them to similarity maps:
\begin{equation}
\label{eq:proto}
S^{n}_{i,l} = \operatorname{Red}_n\Big(1-\cos(\tilde{V}_{i,l}, \boldsymbol{\mu}^{n}_l)\Big), \quad
S^{a}_{i,l} = \operatorname{Red}_a\Big(\cos(\tilde{V}_{i,l}, \boldsymbol{\mu}^{a}_l)\Big),
\end{equation}
where \(\operatorname{Red}_n,\operatorname{Red}_a\) denote reductions over prototypes (e.g., \(\max\), \(\min\), or \(\text{mean}\)). We reshape and upsample $S^{n}_{i,l}$ and $S^{a}_{i,l}$ to the image resolution, yielding support-based predictions $P^{n}_{i,l}\in\mathbb{R}^{H\times W}$ and $P^{a}_{i,l}\in\mathbb{R}^{H\times W}$, which quantify deviation from normality and affinity to known abnormal patterns, respectively.
Finally, we fuse the text-based prediction with the support-based cues at each stage, and then aggregate the fused predictions across all stages to obtain the final map $P_{i}$:
\begin{equation}
\label{eq:fuse}
P^{\text{final}}_{i,l}= \frac{1}{3}\Big(P^{\text{text}}_{i,l}+P^{n}_{i,l}+P^{a}_{i,l}\Big),
\qquad
P_i=\sum_{l=1}^{4} P^{\text{final}}_{i,l}.
\end{equation}
This design effectively combines the generalization capability of CLIP's text alignment with the specific instance-level knowledge from the support set.

\section{Experimental Results}
\textbf{Datasets and Evaluation Metrics.} We evaluate our proposed method on three publicly available datasets with segmentation annotations from a commonly used benchmark BMAD~\cite{bao2024bmad}, spanning three modalities and three different lesion types: LiverCT~\cite{bilic2023liver,landman2015miccai} for liver tumor detection with CT images, RESC~\cite{hu2019automated} for retinal edema detection with OCT images, and BrainMRI~\cite{baid2021rsna,bakas2017advancing,menze2014multimodal} for brain tumor detection with MRI images.
On each dataset, four different shots are considered for few-shot evaluation, including $k=2,4,8,16$.
The Dice score and the area under the Receiver Operating Characteristic curve (AUC) are used to evaluate the anomaly segmentation and classification performance, respectively.

\noindent \textbf{Implementation Details.}
Following~\cite{huang2024adapting,shiri2025madclip}, we adopt the CLIP model with a ViT-L/14 backbone. The vision encoder's 24 transformer layers are evenly split into four stages of six layers each. As in~\cite{huang2024adapting}, we insert a learnable adapter at the end of each vision encoder stage, keeping all other CLIP components frozen. DINOv3 weights are also frozen.
The input image size is 512, with a crop size of 336 and a 2/3 sliding overlap in our SWA scheme.
An Adam optimizer with a batch size of 1 and an epoch number of 50 is adopted for training on an NVIDIA A40 GPU.
The residual connection coefficient $\lambda$ in Eq.~\eqref{eq:refine} is empirically set as 0.6 for LiverCT and 0.9 for RESC.
For BrainMRI, $\lambda$ is 0.9 for $k=4,8,16$ and 0.995 for $k=2$ to ensure training stability.
We use a constant learning rate of 0.001 for LiverCT. For RESC and BrainMRI, a cosine decay schedule is applied, starting from 0.0005 for RESC. For BrainMRI, the initial rate is 5e-5 for 2-shot and 5e-4 for other settings.
During inference, $n_n$ is 50 for all datasets.
$n_a$ is set to 4 (LiverCT), 4 (RESC), and 5 (BrainMRI).
The reduction operation for $\operatorname{Red}_n$ is the minimum across all datasets.
For $\operatorname{Red}_a$, we use the maximum on LiverCT and the mean operation for the other two datasets.

\begin{figure}[!htbp]
 \centering
 \includegraphics[width=0.95\textwidth]{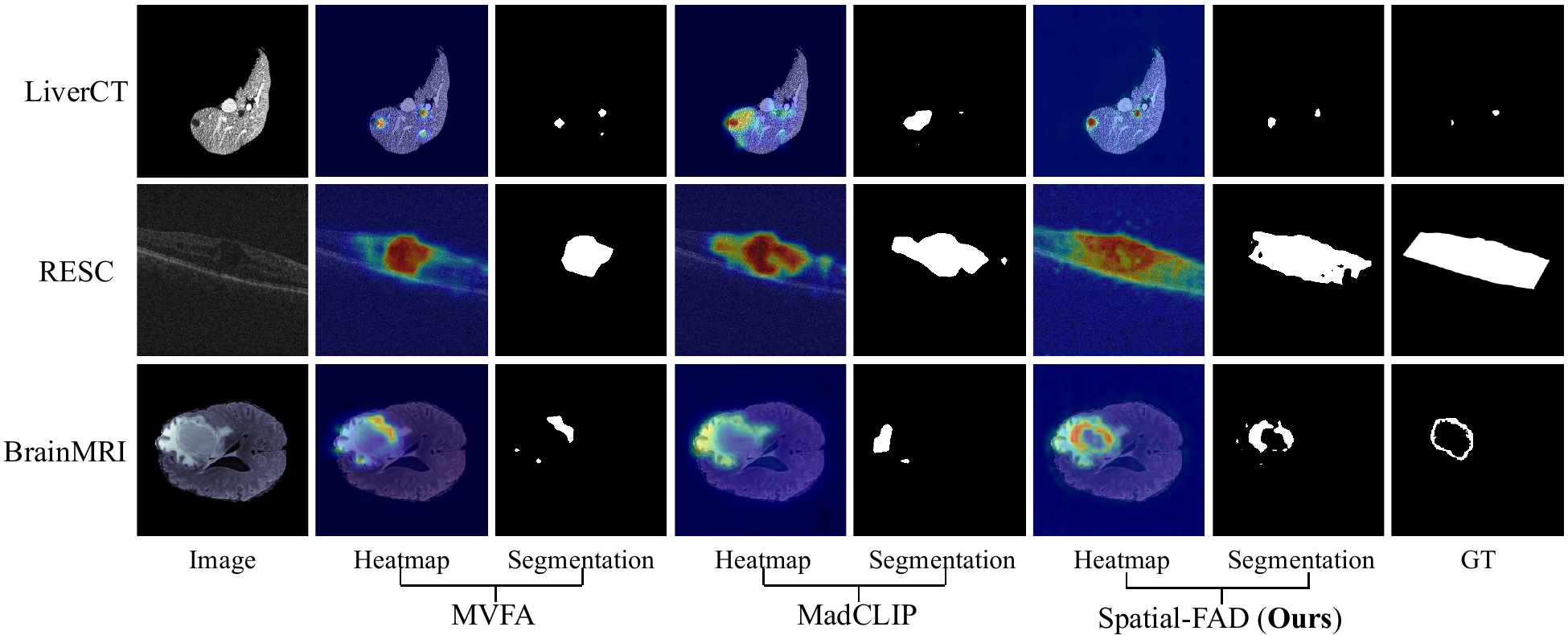}
 \caption{Visualizations of different methods under the setting of 8-shot.}
 \label{visual}
\end{figure}

\noindent \textbf{Comparisons with State-of-the-Arts.} We compare our method with SOTA few-shot anomaly detection methods, including BGAD~\cite{yao2023explicit}, MediCLIP~\cite{zhang2024mediclip}, MVFA~\cite{huang2024adapting}, and MadCLIP~\cite{shiri2025madclip}.
As shown in Table~\ref{compare}, thanks to the effective integration of VFM-induced spatial priors, sliding-window aggregation scheme, and prototype-enhanced support memory, our proposed method achieves the best performance in most cases and obtains significantly superior average results on three datasets across four shot numbers.
Specifically, our method outperforms other SOTA methods by at least 19.76\%, 23.73\%, 21.85\%, and 21.86\% in terms of Dice on the LiverCT dataset under the settings of 2, 4, 8, and 16 shots, respectively.
It is also notable that with only 4-shot examples on the LiverCT dataset, our method obtains a higher Dice and AUC than other SOTA methods trained with 8-shot and even 16-shot data.
On the RESC and BrainMRI datasets, our method also obtains the best Dice in almost all cases.
As indicated by the last two columns of Table~\ref{compare}, our proposed method obtains a Dice improvement of more than 8.9\% on the average of three datasets across various shot numbers.
The AUC score of our method is also superior to others by over 1.8\%, with 4 and 16 shots examples.
The visualization results in Fig.~\ref{visual} also illustrate the superiority of our proposed method with much higher segmentation accuracy.

\begin{table}[!t]
\caption{{Comparisons with SOTA methods on three datasets. The best results are highlighted in bold, while the second best are underlined.}}
\label{compare}
\centering
\scriptsize
{
\begin{tabular}{c|l|cc|cc|cc|cc}
\hline
\multirow{2}{*}{\#Shots} & \multicolumn{1}{c|}{\multirow{2}{*}{Method}} & \multicolumn{2}{c|}{LiverCT} & \multicolumn{2}{c|}{RESC} & \multicolumn{2}{c|}{BrainMRI} & \multicolumn{2}{c}{Average} \\ \cline{3-10} 
                     & \multicolumn{1}{c|}{}                        & Dice↑          & AUC↑          & Dice↑        & AUC↑         & Dice↑          & AUC↑           & Dice↑         & AUC↑          \\ \hline
\multirow{5}{*}{2}   & BGAD (CVPR'23)~\cite{yao2023explicit}                               &  -             &  72.27            & -            &  83.58           & -              &   78.70            & -             &  78.18            \\
                     & MediCLIP (MICCAI'24)~\cite{zhang2024mediclip}                         & 9.77          & 80.42        & \underline{61.14}       & 86.20       & {16.36}         & 92.62         & 29.09        & 86.41        \\
                     & MVFA (CVPR'24)~\cite{huang2024adapting}                               & 16.36         & 81.08        & 47.73       & 91.36       & 2.81          & \underline{92.72}         & 22.30        & 88.39        \\
                     & MadCLIP (MICCAI'25)~\cite{shiri2025madclip}                          & \underline{16.96}         & \underline{84.48}        & 58.55       & \textbf{95.09}       & \underline{18.49}         & \textbf{93.93}         & \underline{31.33}        & \underline{91.17}        \\
                     & Spatial-FAD (Ours)                                         & \textbf{36.72}         & \textbf{86.94}        & \textbf{66.55}       & \underline{94.72}       & \textbf{26.32}              &  {92.51}             & \textbf{43.20}             & \textbf{91.39}              \\ \hline
\multirow{5}{*}{4}   & BGAD (CVPR'23)~\cite{yao2023explicit}                               &  -             &  72.48            &   -          &  86.22           &  -             & 83.56              &   -           & 80.75             \\
                     & MediCLIP (MICCAI'24)~\cite{zhang2024mediclip}                         & 9.00          & 76.48        & 61.81       & 89.37       & 17.92         & 88.43         & 29.58        & 84.76        \\
                     & MVFA (CVPR'24)~\cite{huang2024adapting}                               & 31.18         & 81.18        & \textbf{80.47}       & \underline{96.18}       & 21.03         & 92.44         & \underline{44.23}        & 89.93        \\
                     & MadCLIP (MICCAI'25)~\cite{shiri2025madclip}                          & \underline{32.60}         & \underline{82.97}        & 59.30       & \textbf{96.62}       & \underline{23.21}         & \textbf{95.95}         & 38.37        & \underline{91.85}        \\
                     & Spatial-FAD (Ours)                                         & \textbf{56.33}         & \textbf{92.39}        & \underline{77.24}       & 94.72       & \textbf{33.37}         & \underline{93.93}         & \textbf{55.65}        & \textbf{93.68}        \\ \hline
\multirow{5}{*}{8}   & BGAD (CVPR'23)~\cite{yao2023explicit}                               &  -             & 74.60             &  -           &  89.96           &     -          & 88.01              &   -           &  84.86            \\
                     & MediCLIP (MICCAI'24)~\cite{zhang2024mediclip}                         & 7.35          & 82.04        & 63.69       & 88.30       & 18.83         & 90.77         & 29.96        & 87.04        \\
                     & MVFA (CVPR'24)~\cite{huang2024adapting}                               & 38.04         & 85.90        & 70.07       & 96.57       & 28.29         & \underline{92.61}         & 45.47        & 91.69        \\
                     & MadCLIP (MICCAI'25)~\cite{shiri2025madclip}                          & \underline{38.57}         & \underline{89.31}        & \underline{74.71}       & \underline{97.16}       & \underline{30.36}         & \textbf{95.17}         & \underline{47.88}        & \underline{93.88}        \\
                     & Spatial-FAD (Ours)                                         & \textbf{60.42}         & \textbf{94.69}        & \textbf{75.70}       & \textbf{97.23}       & \textbf{34.41}         & 91.45         & \textbf{56.84}        & \textbf{94.46}        \\ \hline
\multirow{5}{*}{16}  & BGAD (CVPR'23)~\cite{yao2023explicit}                               &  -             &  78.79            &  -           &   91.29          &   -            &  88.05             &     -         & 86.04             \\
                     & MediCLIP (MICCAI'24)~\cite{zhang2024mediclip}                         & 9.22          & 84.11        & 61.40       & 87.67       & 19.51         & 92.03         & 30.04        & 87.94        \\
                     & MVFA (CVPR'24)~\cite{huang2024adapting}                               & 20.81         & 83.85        & 80.42       & \textbf{97.25}       & 28.37         & 94.40         & 43.20        & 91.83        \\
                     & MadCLIP (MICCAI'25)~\cite{shiri2025madclip}                          & \underline{39.53}         & \underline{91.46}        & \underline{82.63}       & 94.08       & \underline{33.90}         & \textbf{95.90}         & \underline{52.02}        & \underline{93.81}        \\
                     & Spatial-FAD (Ours)                                         & \textbf{61.39}         & \textbf{95.59}        & \textbf{84.84}       & \underline{96.83}       & \textbf{39.73}         & \underline{95.43}         & \textbf{61.99}        & \textbf{95.95}        \\ \hline
\end{tabular}}
\end{table}

\begin{figure}[!t]
 \centering
 \includegraphics[width=0.95\textwidth]{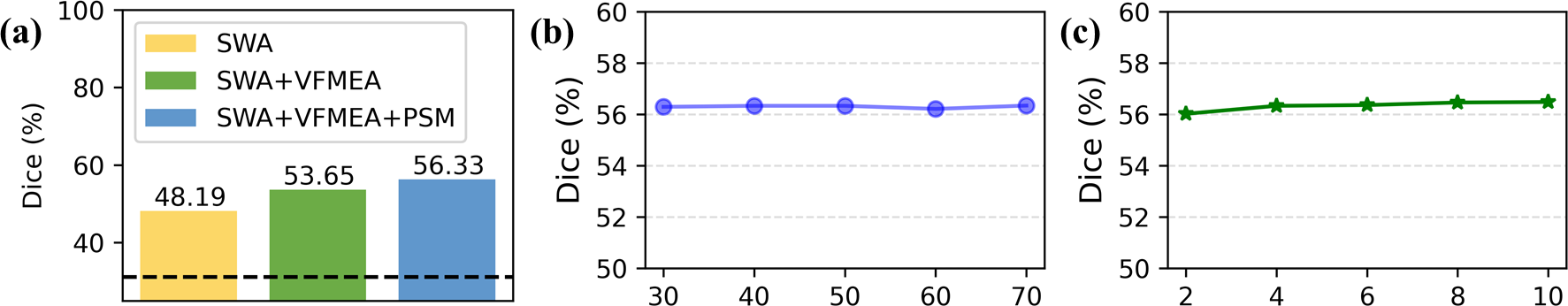}
 \caption{Ablation studies on the LiverCT dataset. (a) shows the effectiveness of each component in our proposed method. {(b) and (c) explore the effects of the number of normal and abnormal prototypes, respectively.}}
 \label{fig:ablation}
\end{figure}

\noindent \textbf{Ablation Studies.} Fig.~\ref{fig:ablation} (a) shows ablation studies on the key components of our method on the LiverCT dataset under the 4-shot setting.
{Compared to the baseline indicated by the dashed line, the sliding-window aggregation strategy can effectively improve segmentation performance thanks to the improved feature resolution.
Furthermore, the core methodological contributions (VFMEA and PSM) are highly complementary to SWA to further enhance spatial resolution and refine the final localization quality by incorporating fine-grained local structural information and maximizing the use of limited support data. In specific, by}
integrating the strong structural affinity priors provided by DINOv3, the Dice score is further improved by 5.46\%.
Finally, the PSM test-time strategy can bring another 2.68\% gain in Dice by fully leveraging the information contained in few-shot training examples.
We also conduct a detailed analysis of the effects of the number of prototypes for normal and abnormal images in the PSM scheme.
Results presented in Fig.~\ref{fig:ablation} (b) and (c) demonstrate that the proposed memory scheme is not sensitive to the number of prototypes, improving the practicality of the method.
Nonetheless, it is interesting to see that the memory bank for abnormal images prefers a small number of prototypes.
Increasing the prototype number to 50 for abnormal images would lead to a Dice decrease of 0.67\%.
This may be because lesions typically have some common features, such as dark areas on CT images for liver tumors, rather than very diverse and extensive characteristics.

\section{Conclusion}

This work pinpoints a key bottleneck in CLIP-based few-shot medical anomaly detection, i.e., strong image-text alignment but weak spatial grounding.
To enhance the spatial accuracy of anomaly detection, we propose Spatial-FAD to integrate VFM spatial priors into CLIP, and further improve localization with sliding-window aggregation and a lightweight prototype-based support memory scheme.
{Results on three datasets show consistent gains in lesion segmentation, achieving an 11.4$\%$ average Dice improvement in the 4-shot setting, together with competitive average AUC performance.}

\begin{credits}
\subsubsection{\ackname} The work described in this paper was supported by the Research Grants Council of the Hong Kong Special Administrative Region, China, under Project T45-401/22-N.

\subsubsection{\discintname}
The authors have no competing interests to declare.
\end{credits}

    

%
%
%
%
\bibliographystyle{splncs04}
\bibliography{Paper-3052}
\end{document}